\documentclass[journal]{IEEEtran}
\ifCLASSINFOpdf
\else
\fi

\usepackage{graphicx}

\usepackage{amsmath}
\usepackage{multirow}
\usepackage{soul}
\usepackage{subfigure}

\setul{0.2ex}{0.15ex}

\usepackage{amsfonts,amssymb}
\usepackage{booktabs}
\usepackage{graphicx}
\usepackage{bbding}
\usepackage{pifont}
\usepackage{wasysym}
\usepackage{makecell}
\usepackage{multicol}

\begin{document}
%
\title{DDRF: Denoising Diffusion Model for Remote Sensing Image Fusion}
%
%
%

\DeclareRobustCommand*{\authorrefmarknumber}[1]{%
    \raisebox{0pt}[0pt][0pt]{\textsuperscript{\footnotesize\ensuremath{#1}}}}

\author{
\IEEEauthorblockN{
ZiHan~Cao{\IEEEauthorrefmark{2}\authorrefmarknumber{1}\thanks{\IEEEauthorrefmark{2} Equal contribution to this work.}},
ShiQi~Cao{\IEEEauthorrefmark{2}\authorrefmarknumber{2}},
Xiao~Wu{\authorrefmarknumber{1}},
JunMing~Hou{\authorrefmarknumber{3}},
Ran~Ran{\authorrefmarknumber{1}}, and
LiangJian~Deng{\IEEEauthorrefmark{1}\authorrefmarknumber{1},~\IEEEmembership{Member,~IEEE}\thanks{\IEEEauthorrefmark{1} Corresponding author.}}
}
}

\markboth{Journal of \LaTeX\ Class Files,~Vol.~13, No.~9, September~2014}%
{Shell \MakeLowercase{\textit{et al.}}: Bare Demo of IEEEtran.cls for Journals}
%



\maketitle

\begin{abstract}
Denosing diffusion model, as a generative model, has received a lot of attention in the field of image generation recently, thanks to its powerful generation capability. However, diffusion models have not yet received sufficient research in the field of image fusion. In this article, we introduce diffusion model to the image fusion field, treating the image fusion task as image-to-image translation and designing two different conditional injection modulation modules (i.e., style transfer modulation and wavelet modulation) to inject coarse-grained style information and fine-grained high-frequency and low-frequency information into the diffusion UNet, thereby generating fused images. In addition, we also discussed the residual learning and the selection of training objectives of the diffusion model in the image fusion task. Extensive experimental results based on quantitative and qualitative assessments compared with benchmarks demonstrates state-of-the-art results and good generalization performance in image fusion tasks. Finally, it is hoped that our method can inspire other works and gain insight into this field to better apply the diffusion model to image fusion tasks. Code shall be released for better reproducibility.
\end{abstract}




%
\IEEEpeerreviewmaketitle

\section{Introduction}
\IEEEPARstart{P}ansharpening, as a fundamental problem in remote sensing image fusion processing, it attracts more interest from research community and commercial companies. Pansharpening requires a high spatial resolution panchromatic (PAN) image and a low spatial resolution multispectral (LrMS) image, and fuse two domains together to obtain a high spatial resolution multispectral image (HRMS) which preserves the advantages of the two domains. Most satellites can simultaneously capture PAN and MS images, such as World-View3 and GaoFen2. Pansharpening can also be seen as a preprocessing method for further high-level applications, such as change detection~\cite{wu2017post, bandara2022ddpm},  mineral exploration~\cite{bishop2011hyperspectral, carrino2018hyperspectral} and segmentation~\cite{hossain2019segmentation}.

Pansharpening methods are usually divided into four categories, i.e., component substitution (CS) methods, variational optimized (VO) techniques, multiresolution analysis (MRA) methods and machine learning (ML) approaches. The CS method substitute a component of the LrMS image by the PAN image, contains algorithms such as principal component analysis (PCA)~\cite{kwarteng1989extracting}, gram-schmidt spectral sharpening~\cite{laben2000process, aiazzi2007improving} and guided filtering~\cite{qu2017hyperspectral}. The CS-based methods can generate a fused image with high spatial fidelity but more spectral distortion. The MRA approach extracts features on the spatial dimension from PAN image and then injects them into LrMS image in a multispectral manner. The representative MRA methods are the addictive wavelet luminance proportional (AWLP)~\cite{otazu2005introduction}, smoothing filter-based intensity modulation (SFIM)~\cite{liu2000smoothing}, modulation transfer function (MTF)~\cite{aiazzi2006mtf} generalized laplacian pyramid (MTF-GLP) and MTF-GLP with high-pass modulation (MTF-GLP-HPM)~\cite{vivone2013contrast}. The MRA method preserves better spectral information but suffers from spatial distortion due to its extract-injection manner. The VO-based approaches are recently proposed to address pansharpening issue. Algorithms belonging to this class include Bayesian methods~\cite{he2014new, wang2018high}, variational methods~\cite{moeller2009variational, duran2014nonlocal, xu2020hyperspectral}. VO-based methods fuse better HRMS images owing to their mathematical rigor with respect to the state-of-the-art CS-based and MRA-based methods. However, the performance improvement brings higher computational costs and more tunable parameters.

Over recent years, with the improvement of the hardware and deep learning, convolutional neural networks (CNN)~\cite{resnet, unet} and vision Transformers (ViT)~\cite{swin, pvt} have been proposed and achieved significant advances in image processing, e.g. image classification~\cite{resnet,swin}, image denosing~\cite{lehtinen2018noise2noise}, image deblur~\cite{NAFNet}, image super-resolution~\cite{NAFNet++} and image fusion~\cite{deng2020detail, deng2023psrt, xu2020u2fusion}. Therefore, those DL-based methods quickly become a powerful tool to handle pansharpening task by giving large-scale training datasets and learn a nonlinear functional mapping from LrMS and PAN to HRMS. PNN~\cite{pnn}, as a simple but effective CNN, is firstly proposed to deal with pansharpening task. This architecture is inspired by previous CNN for single image super-resolution~\cite{dong2015image}. PanNet~\cite{yang2017pannet} is proposed to explicitly inject high-passed information from PAN image into the upsampled LrMS image and gains a better performance. He \textit{et al.}~\cite{dicnn} proposed a detail injection-based CNN. Deng \textit{et al.}~\cite{deng2020detail} employs a residual network by stacking several residual blocks and another residual path from input to output to let the network learn high-frequency details. Those CNN-based methods are constricted by static kernels, which means the kernels are independent of the input. To alleviate this drawback, a dynamic kernel-generated module in LAGNet~\cite{jin2022lagconv} is inserted to generate dynamic kernels depending on the input.

\begin{figure}
    \centering
    \includegraphics[width=1.0\linewidth]{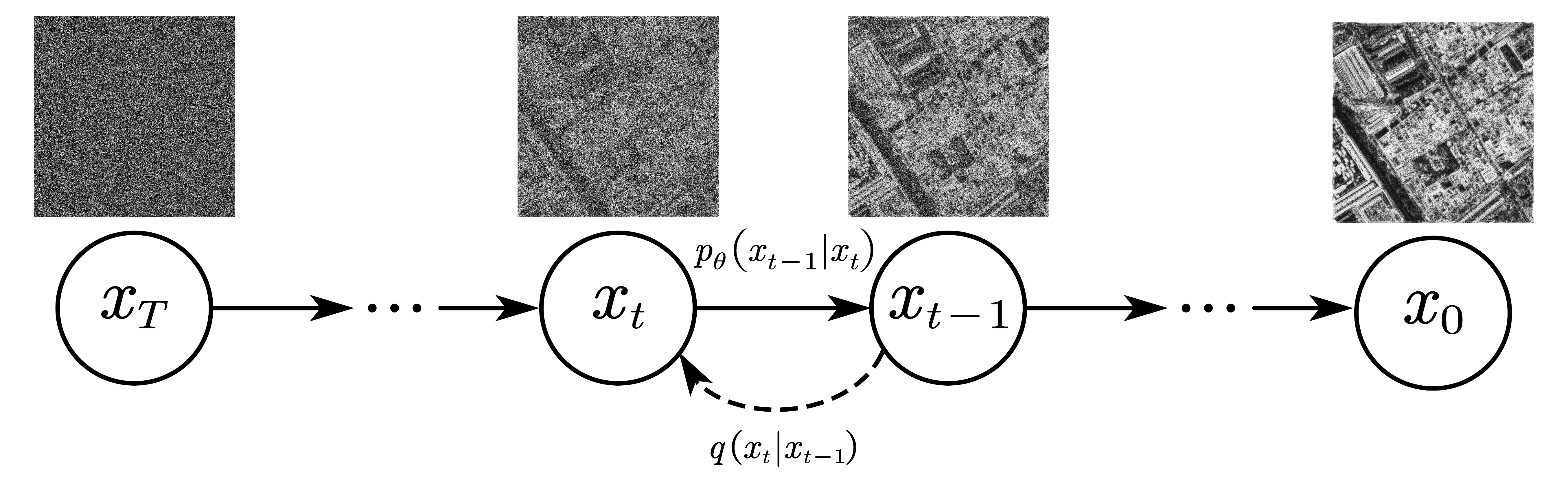}
    \caption{Denoising diffusion model forward and backward process. $q(\mathbf x_t|\mathbf x_{t-1})$, $p_\theta(\mathbf x_{t-1}|\mathbf x_t)$ represent noising adding forward process and denoising backward process, respectively.}
    \label{fig:forward_backward_process}
\end{figure}

Diffusion denoising probability model (DPM), as proposed for unconditional image generation~\cite{bandara2022ddpm, improved_ddpm}, conditional image generation~\cite{bandara2022ddpm, karras2022elucidating}, text-to-image generation~\cite{imagen, stable_diffusion}, image-to-image translation~\cite{saharia2022palette} and discrete nature language generation~\cite{diffusion_lm} tasks, has shown its power providing extra feature details and good generation ability. Another advantage is that DPM owns a more stable training process and merely no mode collapse compared to the GAN-based model. The training and testing phase of DPM is dubbed as forward process and backward process~\cite{song2020score, ncsn}. In the forward process, the input image is corrupted by pure Gaussian noise and the model tries to remove the added noise and recover the original input. In the backward process, the input is pure Gaussian noise and the trained model is responsible for removing a little noise step by step, so, after large enough time steps, the input pure noise is recovered to the generated images. To control the generated images, Ho \textit{et al.}~\cite{classfier-free-guidance} propose conditional DPM, the condition is fed into the diffusion model to control the SDE~\cite{bandara2022ddpm, song2020score, ncsn} or ODE~\cite{ddim, dpm_solver} trajectory.

In this paper, We propose DDRF for pansharpening. Mainstream works on pansharpening often concentrate on model architectures and information gathering operations. Most of them choose to fuse two domains directly and few notice generative models can hallucinate more detailed information as they are trained in a generative manner. Those non-generative models often suffer high-frequency component loss, spectral or spatial distortions. On the contrary, DPM with SDE sampling can introduce new noise that contains many high-frequency components when sampling for better generation. Our contribution can be summarised in the following four folds:
\begin{enumerate}
    \item We firstly propose DDRF for pansharpening, which is a supervised diffusion model but modified to suit fusion tasks and can be also extended to other different modal fusion tasks (e.g. hyperspectral image fusion). While achieving good performance, we hope our DDRF can draw attention to diffusion models in the field of image fusion.
    \item Two conditional injection modules are designed that are used for injecting conditions in a style transfer manner and a details complemented manner, respectively.
    \item Our DDRF reaches a new state-of-the-art performance on pansharpening task among three widely used pansharpening datasets and competitive performance on hyperspectral dataset.
\end{enumerate}

\section{Related works}
This section mainly introduces generative and non-generative methods related to pansharpening, as well as mainstream generative tasks and methods related to diffusion models.
\subsection{Non-generative Model}
PNN~\cite{pnn} was inspired by the single image super-resolution technique and proposed a convolutional neural network for pansharpening for the first time. PanNet~\cite{yang2017pannet} is proposed to explicitly inject high-passed information from the PAN image into the
upsampled LRMS image and gain a better performance. DiCNN~\cite{dicnn} is proposed as a details-inject-based CNN which is powerful on frequency details. FusionNet~\cite{deng2020detail} simply employs a residual from the start to the end of the network, explicitly enabling the network to learn the high-frequency details.
Due to the static kernel of the conventional convolutional operator, LAGNet~\cite{jin2022lagconv} with dynamic kernel generated based on input can better fuse two domains. AKD~\cite{akd} further exploits the idea of LAGNet and proposes two dynamically generated branches responsible for spectral and spatial details extraction, respectively. SpanConv~\cite{chen2022spanconv} presents an interpretable span strategy, for the effective construction of kernel space and reducing the redundancy of the convolution but still maintaining good performance. PMACNet~\cite{liang2022pmacnet}, a parallel convolutional neural network structure, is employed with a pixel-wise attention constraint (PAC) module and a novel multireceptive-field attention block (MRFAB), successfully solved the small receptive field brought by CNN. 

\subsection{Generative Model}
Generative models, as another promising branch for multi-modal fusion, including GANs, flow-based models, energy-based models, and diffusion models have been proposed from different perspectives and have received a lot of attention and research, leading to the development of various types of generative tasks. Benefiting from the development of these generative models, generative models can also be applied to pansharpening, producing in good results.

\textbf{GAN-based model} is first introduced by Goodfellow \textit{et al.}~\cite{gan} including a generative model $G$ that generates a sample on the data distribution, and a discriminative model $D$ that estimates the probability that a sample comes from the training data rather than $G$. The training procedure for $G$ is to maximize the probability of $D$ making a mistake. Through this minmax two-player adversarial training procedure, the generative ability of $G$ is promoted. To deal with the training instability and failure to converge, W-GAN~\cite{w-gan} is presented using the Wasserstein distance as the loss function, which provides more accurate gradient information and avoids the gradient vanishing and mode collapse issues during training. Improved W-GAN~\cite{improved-wgan} further provides an alternative to clipping weights: penalize the norm of the gradient of the critic with respect to its input and achieve higher-quality generations. StyleGAN v1~\cite{stylegan1} and StyleGAN v2~\cite{Karras2019stylegan2} further push forward the stability and diversity of GAN.

Ma \textit{et al.}~\cite{pan_gan} firstly introduced GANs into pansharpening and performed unsupervised learning of spectral and spatial information based on the reversed loss function upon the network's output. Zero-GAN~\cite{zero_gan_ieee} also relies on GAN and constructed a non-reference loss function, including an adversarial loss, spatial and spectral reconstruction losses, a spatial enhancement loss, and an average constancy loss to specifically supervise the learning process.

\textbf{Flow-based model} is a type of generative model that learns the underlying distribution of data by transforming a simple input distribution into the target distribution through a series of invertible transformations. This is achieved by representing the probability density function of the target distribution as a sequence of flows, where each flow is an invertible transformation.
Some notable contributions in the field of flow-based models include the Real-valued Non-Volume-Preserving (Real NVP) model, introduced by Dinh \textit{et al}~\cite{real_nvp}, which proposed a flexible and efficient way to construct invertible transformations using neural networks. Another important contribution is Glow~\cite{kingma2018glow}, which improved upon previous flow-based models by introducing an invertible 1$\times$1 convolutional operation that greatly increased the expressiveness of the model.


\textbf{Diffusion model} are recently proposed for generation including conditional or unconditional generation~\cite{bandara2022ddpm, improved_ddpm, song2020score}, text-to-image translation~\cite{stable_diffusion, diffae}, image super-resolution~\cite{saharia2022image}, image restoration~\cite{ddrm, ddnm} and other high-level image manipulation tasks~\cite{wang2022sindiffusion, zhang2023layoutdiffusion, saharia2022palette}. Diffusion models can generate images with more details and higher fidelity than the GAN-based model and flow-based model. The advantages of diffusion models include stable training, minimal mode collapse, and the ability to train with only a single mean squared error (MSE) loss. In comparison to GAN-based models, which suffer from instability issues in their adversarial training, and flow-based models, which are limited in their network performance due to the requirement for reversibility, diffusion models are easier to train and design.

Song \textit{et al.}~\cite{SMLD_yangsong} first introduced a score-based model that produces samples via Langevin dynamics using gradients of the data distribution estimated with score matching. Later, Ho \textit{et al.}~\cite{bandara2022ddpm} proposed DDPM from the direction of weighted variational bound, and their equivalence is proven in NCSN~\cite{ncsn}. DDIM~\cite{ddim} designed a non-Markov chain sampling process, which accelerated the sampling of diffusion models. DPM-solver~\cite{dpm_solver} simplified the solution to an exponentially weighted integral of the neural network by computing the linear part of the ODE and applying change-of-variable, further accelerating the sampling process. EDM~\cite{karras2022elucidating} decoupled various design components of the diffusion model and designed a second-order ODE sampler, which further improved the performance of the diffusion model to reach the state-of-the-art performance.

However, diffusion models have not received much attention in remote sensing images. DDPM-CD~\cite{ddpm-cd} utilizes diffusion models for landform change detection. Firstly, an unconditional diffusion model is trained on a large dataset, and the features of a specific layer of the diffusion model during the sampling process are used as additional information input to the segmentation head, producing segmentation results. As a result, DDPM-CD has achieved satisfactory performance in the field of landform change detection. Similar to DDPM-CD, Dif-fuse~\cite{dif-fuse} fuses RGB and near-infrared images and has made considerable progress.

\section{Methodology}
Our DDRF will be elucidated in the following parts. Firstly, some notations will be explained. Then diffusion denoising model will be briefly reviewed, including the forward and backward processes, and the simple derivation of the loss function. Subsequently, we will provide a detailed introduction to the architecture of the diffusion model and two types of conditional injection modules. After that, according to the training objective of the diffusion model, we discuss the three choices (i.e., $\epsilon$, $\mathbf x_0$, and $\mathbf v$). To facilitate faster sampling, we will also discuss how to transform the VP SDE into a faster-sampling ODE while preserving the marginal distribution in the final subsection. 

\subsection{Notation}
We first clarify the notations. An image of size $\mathbb R^{H,W,C}$ represented by a tensor is denoted as $\mathbf x$. After performing $t$ forward diffusion steps, the resulting noisy image is denoted as $\mathbf x_t$. In particular, we denote the PAN image and the upsampled LrMS image of the same size as the PAN image as $\mathbf P \in \mathbb R^{H\times W \times 1}$ and $\mathbf M \in \mathbb R^{H\times W\times C}$, respectively. The feature representation of the intermediate layers of a neural network is denoted as $\mathbf f$. The condition that guides the generation process is represented as $\mathbf c$.

\subsection{Diffusion Model}
Diffusion models enable generating a realistic image from a normal distribution by reversing a gradual noising process. The diffusion model accomplishes its task in two steps: forward and reverse processes which are illustrated in Fig. \ref{fig:forward_backward_process}.

Forward process aims to make the origin image $\mathbf{x_0}\sim p_{data}(\mathbf {x_0})$ noisy by a $T$ step Markov chain that gradually converts it to Gaussian distribution. One forward step is defined as,
\begin{equation} \label{eq: one_forward_step}
    q(\mathbf x_t|\mathbf x_{t-1})=\mathcal N(\mathbf x_t; \sqrt{1-\beta_t} \mathbf x_{t-1},\beta_t \mathbf I),
\end{equation}
where $t\in [0, T]$ and $\mathcal N$ is a Gaussian distribution which mean is $\sqrt{1-\beta_t}\mathbf x_{t-1}$ and standard deviation is $\beta_t \mathbf I$.
Through the reparameterization trick, the manner we directly get $x_{t}$ is defined as 
\begin{equation} \label{eq: forward_prob}
q(\mathbf x_{t}|\mathbf x_{0})=\sqrt{\bar{\alpha}_{t}}\mathbf x_{0}+\sqrt{1-\bar{\alpha}_{t}}\epsilon,
\end{equation} 
where $\epsilon \sim \mathcal N(0,1)$ and  $\beta_{t}\in (0, 1)$ is a pre-defined variance schedule in $T$ steps and $\alpha_t=1-\beta_t, \bar \alpha_t=\prod_{i=0}^t \alpha_i$.

    
We do reverse processes to denoise $x_{t}$, following 
\begin{equation} \label{eq: reverse_process}
    p_\theta(\mathbf x_{t-1}|\mathbf x_t)=\mathcal N(\mathbf x_{t-1};\mu_\theta(\mathbf x_t,t),\Sigma_\theta(\mathbf x_t,t)),
\end{equation}
where $\theta$ denotes the parameters of the model.


The forward process degrades the data distribution into a standard Gaussian distribution. On the contrary, the reverse process, modeled by a neural network, aims to learn to remove the degradation brought from the forward process, i.e., denoising.

To train a diffusion model, maximizing its variational lower bound (VLB) and it can turn into a simple supervised loss~\cite{bandara2022ddpm} written as,
\begin{equation} \label{ddpm_loss}
    \min_\theta L_{simple} = \mathbb E[\|\mathbf \epsilon-\epsilon_\theta(\mathbf x_t,t)\|_1],
\end{equation}
Note that in Eq. \ref{ddpm_loss}, the model prediction is added Gaussian noise, but it can also be $\mathbf x_0$~\cite{bandara2022ddpm} or $\mathbf v$~\cite{imagen}. Some discussions will be made in the ablation study.

After training the model, we can sample data starting a standard Gaussian noise $\mathbf x_T$, and according to Eq. \ref{eq: reverse_process}, the mean and variance can be computed, following
\begin{equation}
    \mu_\theta=\cfrac 1{\sqrt{\alpha_t}}(\mathbf x_t-\cfrac{\beta_t}{\sqrt{1-\bar \alpha_t}}\epsilon_\theta(\mathbf x_t,t)),
\end{equation}
\begin{equation} \label{eq: sigma}
    \Sigma_\theta(\mathbf x_t,t)=\cfrac{1-\bar \alpha_{t-1}}{1-\bar \alpha_t}\beta_t.
\end{equation}
To this end, $\mathbf x_{t-1}$ can be sampled from the previous step. Through $T$ iterative sampling steps, $\mathbf x_0$ can be obtained.

In the image fusion task, there usually are two or even more images from different domains. Take pansharpening as an example, PAN image and LrMS images are from different domains and the HRMS image is from the target domain. We treat the fusion task as an image-to-image translation task. To guide the generation process, we empirically set the LrMS, PAN, and constructed wavelet coefficients (present in sec. \ref{sec: wavelet_modulation}) as conditions denoted as $\mathbf c$. Then the forward, the backward distributions and training objective (i.e., Eq. \ref{eq: one_forward_step}-\ref{eq: sigma}) shall be conditioned on $\mathbf{c}$.

\subsection{Deep Model Architecture}
\begin{figure*}
    \centering
    \includegraphics[width=0.8\linewidth]{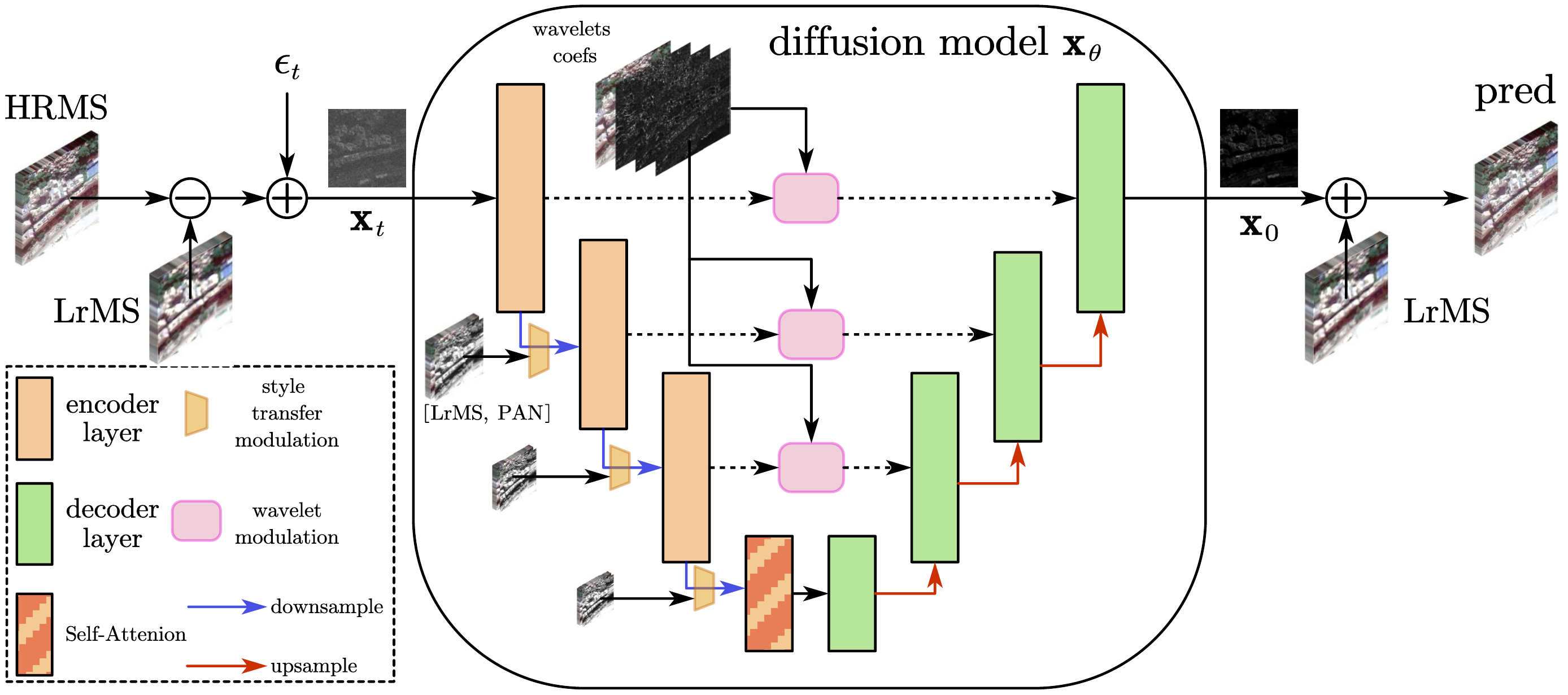}
    \caption{Framework of our DDRF. The input of DDRF is the residual produced by HRMS subtracted by MS. For simplicity, the time step is omitted. Zooming in for more details.}
    \label{fig:arch}
\end{figure*}
As shown in Fig. \ref{fig:arch}, the overall network architecture of our DDRF is similar to SR3~\cite{saharia2022image} and is composed of an encoder, decoder, and two middle self-attention modules, which resembles an Unet architecture consisting of convolutional layers. There are two downsampling operations between the two layers of the encoder to reduce spatial resolution and increase the number of channels. Similarly, there is an upsampling operation by a factor of 2 between the two layers of the decoder to increase spatial resolution and reduce the number of channels.

The encoder and decoder have the same number of layers. Differently, the encoder takes in encoded features not only from the previous layer but modulated coarse-grained style condition (seen in sec. \ref{sec: style_transfer}). The features of the encoder at the corresponding layer will be concatenated with the input features of the decoder at the corresponding layer and addictive wavelet features (seen in sec. \ref{sec: wavelet_modulation}) along the channel dimension and fed into the decoder together. Its output can be modeled as added Gaussian noise, $\mathbf x_0$ or $\mathbf v$, which will be discussed in sec. \ref{training_objective}.

\subsection{Residual Learning}
Due to the tendency of CNNs to learn low-frequency information, they lack representation of high-frequency information. Previous works have attempted to address this issue by filtering the input with high-pass filters and directly incorporating it as residual input, or by designing specialized high-frequency injection modules and supervising them in the frequency domain. We do not want to design complex high-frequency modules that significantly increase the number of parameters in the network, nor do we want to perform complex operations such as loss computation in the frequency domain.

Inspired by FusionNet~\cite{deng2020detail}, we changed the input of the diffusion model during the training process from HRMS to $\mathrm{HRMS-MS}$ and then add noise based on Eq. \ref{eq: forward_prob}.
We found that using residuals as input for HRMS converges faster and produces better results than when HRMS is used as input. Regarding whether to use residual learning or not, discussion and ablation experiments will be shown in sec. \ref{sec: residual_learning}.

\subsection{Style Transfer Modulation} \label{sec: style_transfer}
We concatenate the MS and PAN along the channel dimension and consider it as a conditional input with a style. We inject this conditional input into each layer of the encoder and encode it together with the residual input. We avoid encoding much detailed information in the encoder but reconstructing those in the decoder, to this end, the encoder only needs to consider the overall style, including the approximate information and distribution of both spectral and spatial features.
\begin{figure}[!ht]
    \centering
    \includegraphics[width=1.0\linewidth]{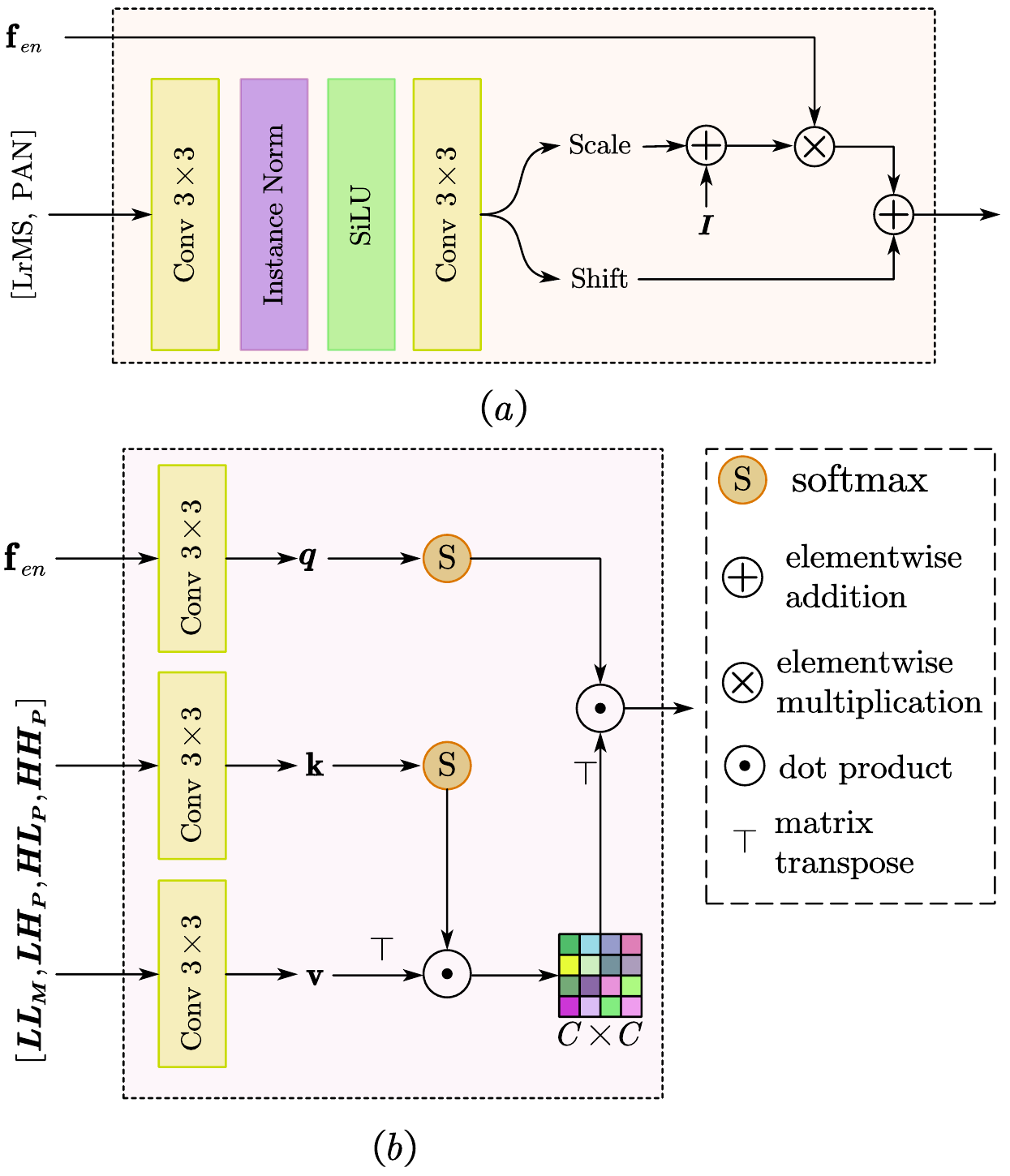}
    \caption{Two types of modulations. (a) style transfer modulation. (b) wavelet modulation.}
    \label{fig: modulation}
\end{figure}

For such coarse-grained operations, it is feasible to generate the corresponding scale and shift based on the input conditions to modulate the output features of the encoder, which can be represented as,
\begin{equation}
\label{style_transfer_condition_injection}
\begin{aligned}
    & \mathrm{scale, shift} = \mathtt{Split}(\text{MLP}([{\bf P, M}])),\\
    & \mathbf{\hat{f}} = \mathbf f\cdot(1+\mathrm{scale})+\mathrm{shift},
\end{aligned}
\end{equation}
where $\mathtt{Split}$ is splitting the feature into two parts with equal size along the channel dimension, $[\cdot]$ represents concatenation along the channel dimension, and ${\bf f} \in \mathbb R^{h,w,d}$ is the output of the previous decoder layer, $\mathbf{\hat f}$ is the modulated output. MLP is implemented as several convolutional layers which are inserted by a $\mathtt{SiLU}$ activation in between. $\mathrm{scale} \in \mathbb R^{h,w,d}$ and $\mathrm{shift} \in \mathbb R^{h,w,d}$ are produced by the MLP. $h, w, d$ are determined by the size of the feature output from the encoder layer.

\subsection{Wavelet modulation} \label{sec: wavelet_modulation}
The encoder layer is responsible for encoding spectral and spatial information into a high dimensional and low spatial resolution feature with treating LrMS and PAN as style conditions. The decoder, its objective is to decode high-dimensional, low spatial-resolution features from the encoder as faithfully as possible into the RGB space, using the features learned by the encoder. However, relying solely on the features outputted by the encoder is not sufficient for the decoder. We have found that introducing appropriate high-frequency components during the decoding process is advantageous for producing images with richer details. One intuitive approach is to concatenate the high-frequency components extracted from PAN with the features outputted by the encoder along the channel dimension, and then feed the concatenated features into the decoder for decoding.  

We choose to apply wavelet decomposition on the PAN to extract the horizontal, vertical, and diagonal high-frequency components and used a cross-attention mechanism in the decoder to introduce these high-frequency details into the decoding process. However, considering only high-frequency details are insufficient, we also introduced the low-frequency main component of LrMS, which is also extracted using wavelet decomposition similar to PAN. The high-frequency and low-frequency components are concatenated in the channel dimension and then passed through an attention module with the features outputted by the encoder before being introduced into the decoder. We refer to this process as \textit{Wavelet Modulation}. The advantage of this approach is that it separates the low-frequency and high-frequency components, with PAN providing the high-frequency details and LrMS providing the spectral component, making it easier for the decoder to learn.

To be more specific, we utilized the DB1 wavelet decomposition to decompose both the LrMS and PAN into four components, which include one main component  and three detail components in the horizontal, vertical, and diagonal  directions.
We can express this process using the following equations,
\begin{equation}
\begin{aligned}
    \mathbf{LL}_{\bf M}, \mathbf{LH}_{\bf M}, \mathbf{HL}_{\bf M}, \mathbf{HH}_{\bf M}
    &=\mathtt{DB1}(\bf M), \\
    \mathbf{LL}_{\bf P}, \mathbf{LH}_{\bf P}, \mathbf{HL}_{\bf P}, \mathbf{HH}_{\bf P}
    &=\mathtt{DB1}(\bf P),
\end{aligned}
\end{equation}
where ${\bf LL}\in \mathbb R^{C, \frac H 2, \frac W 2}$ denotes the low-frequency main component, and ${\bf LH}\in \mathbb R^{C, \frac H 2, \frac W 2}$, ${\bf HL}\in \mathbb R^{C, \frac H 2, \frac W 2}$, and ${\bf HH}\in \mathbb R^{C, \frac H 2, \frac W 2}$ denote the detail components in the horizontal, vertical, and diagonal directions, respectively.

Since the memory complexity of cross-attention is $\mathcal O(n^2)$, where $n=HW$. When approaching the head of the decoder, as the output image size increases, it is unacceptable when dealing with large images due to the quadratic spatial complexity of the image size. we propose a linear-memory cross-attention mechanism to introduce both the high-frequency and low-frequency components into the decoder. We can represent the linear-memory attention mechanism as follows,
\begin{equation}
\begin{aligned}
    &\mathbf{\hat{q}}=\mathtt{Reshape}(\text{Softmax}(\mathbf q, 1)), \\
    &\mathbf{\hat{k}}=\mathtt{Reshape}(\text{Softmax}(\mathbf k, 2)), \\
    &\mathbf{\hat{v}}=\mathtt{Reshape}(\mathbf v), \\
    &\mathbf A=\mathbf{\hat{k}}\cdot \mathbf {\hat v}^\top, \\
    &\mathbf{O}=\mathbf A^\top \cdot \mathbf{\hat q}.
\end{aligned}
\end{equation}
As a result, we reduced the memory complexity to $\mathcal O(C^2)$, where $C$ is much smaller than $n$. This significantly reduces memory usage while ensuring the effective introduction of conditions. $\bf q$ is the output of the corresponding layer in the encoder, and $\bf k$ and $\bf v$ are tensors obtained by concatenating the $\bf LL_M, LH_P, HL_P$, and $\bf HH_P$ components along the channel dimension and then are projected by several convolutional layers as follows,
\begin{equation}
\begin{aligned}
      &\mathbf q = {\bf W_q} \otimes \mathbf f_{en} + b_q,\\
      & [{\bf k, v}] = {\bf W_{k,v}} \otimes [\mathbf{LL_M, LH_P, HL_P, HH_P}] + b_{k,v},
\end{aligned}
\end{equation}
where $\otimes$ is the convolution operation, $\mathbf W$ is the weight and $b$ is bias and $\mathbf{f}_{en}$ is the feature from the corresponding encoder layer. The $\mathtt{Reshape}$ operation combines the spatial dimensions of $\bf q$, $\bf k$, and $\bf v$ into a single dimension, and can reprensent as $\mathtt{Reshape}: \mathbf f\in \mathbb R^{C,H,W}\rightarrow \mathbb R^{C, HW}$. $\text{Softmax}(\cdot, i)$ represents applying the softmax operation along the $i$-th dimension. In this way, the high-frequency details and low-frequency information are injected for better decoding.

\subsection{Training Objective $\mathbf x_0$, $\epsilon$ or $\mathbf v$} \label{training_objective}
There are three choices for the neural network to model, i.e., $\epsilon$, $\mathbf x_0$, and $\mathbf v$~\cite{imagen}. Note that $\mathbf{v}$ is a combination of the $\epsilon$ and $\bf x_0$,
\begin{equation}
    \mathbf v=\sqrt{\bar \alpha_t} \epsilon-\sqrt{1-\bar \alpha_t} \mathbf x_0.
\end{equation}
In previous diffusion works~\cite{bandara2022ddpm, improved_ddpm}, the training objective is often the added Gaussian noise, which is suitable for large-scale datasets in practice. However, remote sensing datasets are much smaller than those nature image datasets. Intuitively, predicting non-degraded images from noisy images is rather more direct than predicting those added noises. We will show that for small datasets, predicting $\mathbf x_0$ is a better choice than noise and $\mathbf v$ in the ablation study.

By turning the objective into $\mathbf x_0$, we can change this $L_{simple}$ to another loss that has the same form but depends on $\mathbf x_0$ instead. The formula is as follows,
\begin{equation}
    L_{simple}=\mathbb E[\|\mathbf x_0-\mathbf x_\theta(\mathbf x_t, t,\mathbf c)\|_1].
\end{equation}

\subsection{Fast Iterative Sampling}
For the backward process, since the diffusion model needs to iterate hundreds or thousands of times to generate an image, this leads to a slow generation speed. To solve this problem, we convert the SDE sampler~\cite{bandara2022ddpm} of the original backward sampling process into an ODE sampler~\cite{ddim}, which allows for fast sampling in a non-Markov chain form as
\begin{equation}
\mathbf x_{t-1} = \sqrt{\bar{\alpha}_{t-1}}\mathbf x_{\theta}(\mathbf x_{t},t)+\sqrt{1-\bar{\alpha}_{t-1}-\sigma_{t}^{2}}\epsilon_{\theta}(\mathbf x_{t},t),
\end{equation}
where $\sigma_{t}$ is a established function about $t$ and $\mathbf x_{\theta}(x_{t},t)$ is the model prediction of $\mathbf x_{0}$, thus we have 
\begin{equation} \label{eq: epsilon2x0}
\mathbf x_{\theta}(\mathbf x_{t},t):=\frac{\mathbf x_{t}-\sqrt{1-\bar{\alpha}_{t}}\epsilon_{\theta}(\mathbf x_{t},t)}{\sqrt{\bar{\alpha}_{t}}},
\end{equation}
we can know from Eq. \ref{eq: epsilon2x0} that the sampler predicts $\mathbf x_{0}$ directly from $\mathbf x_{t}$ and then generates $\mathbf x_{t-1}$ through a reverse conditional distribution. Therefore, we can accelerate sampling by $\tau = [\tau_{1},\tau_{2},...,\tau_{dim(\tau)}]$. Specifically, $\tau$ is a subset of $[1,2,...T]$, and we can make a fast sampling after using $\tau$,
\begin{equation} \label{eq: respacing}
\mathbf x_{\theta}(\mathbf x_{\tau_{i}},\tau_{i}):=\frac{\mathbf x_{\tau_{i}}-\sqrt{1-\bar{\alpha}_{\tau_{i}}}\epsilon_{\theta}(\mathbf x_{\tau_{i}},\tau_{i})}{\sqrt{\bar{\alpha}_{\tau_{i}}}},
\end{equation}
so that the sampling speed can be greatly improved. 
\section{Experiments}

\subsection{Experimental Settings}
\subsubsection{Implementation Details}
our DDRF is implemented in PyTorch 1.13.1 and Python 3.10.9 using AdamW optimizer with a learning rate of 0.0001 to minimize $L_{simple}$
on a Linux operating system with two NVIDIA GeForce RTX4090 GPUs. The initialization of convolution modules is based on a kaiming initialization. To obtain wavelet coefficients, we use $\mathtt{DB1}$ wavelets decomposition method and decompose the image into four wavelet coefficients. For diffusion denoising model, we choose a cosine schedule~\cite{improved_ddpm} for $\alpha_t$ following,
\begin{equation}
    \bar \alpha_t=\frac{f(t)}{f(0)}, f(t)=\cos\left(\frac{t/T+s}{1+s}\cdot \frac \pi 2\right).
\end{equation}
In practice, we set $s=8e-3$. The model does not learn the variance term $\Sigma_\theta$ introduced in improved-ddpm~\cite{improved_ddpm}. The total training diffusion time step is set to 500 for both pansharpening and hyperspectral fusion experiments. The exponential moving average (EMA) ratio is set to 0.995.
The total training iterations for WV3, GF2, QB, and CAVE datasets are set to 100k, 100k, 200k, and 300k iterations, respectively.
To reduce the number of sampling steps, according to Eq. \ref{eq: epsilon2x0} and \ref{eq: respacing}, we set the number of sampling steps to 25 for the pansharpening task, and 100 for the hyperspectral image fusion task.


\subsubsection{Datasets}
In order to show the effectiveness of our DDRF,  we conduct experiments over the widely-used datasets including World-View3 (8 bands),  Gao-Fen2 (4 bands), and Quick-Bird (4 bands) datasets.

For \textbf{pansharpening datasets}, to best evaluate the performances, we simulate the reduced resolution and full resolution datasets to obtain the reference metrics and non-reference metrics.
For the reduced datasets, to obtain the simulated images with ground-truth images, we follow Wald's protocal~\cite{wald1997fusion} to get them. It has three steps: 1) Downsample the original PAN and the original MS images by a factor of 4 using modulation transfer (MTF)-based filters. The downsampled PAN and MS are treated as the training PAN and training MS images; 2) Regard the original MS image as the ground-truth image (i.e., HRMS); and then 3) Upsample the training MS image by interpolating (polynomial kernel) with 23 coefficients~\cite{exp} and treating it as LrMS image. When dealing with the full datasets, the original MS images are considered as MS and the upsampled MS images are treated as LrMS, the original PAN images are seen as PAN.

For \textbf{hyperspectral dataset}, we choose CAVE indoor dataset to further evaluate our DDRF. The CAVE dataset contains 32 hyperspectral images (HSIs) and corresponding multispectral images (MSIs) with a size of 512$\times 512 \times 31$. 20 images are selected for training and validation, and 11 images remained for testing.

\subsubsection{Data Simulation}
For pansharpening datasets, we directly crop the MS image, LrMS image, HRMS image, and PAN image into patches. The train/validation/test split for WV3, GF2, and QB reduced datasets is 9714/1080/20, 19809/2201/20, and 17139/1905/20, respectively. The resolutions of the reduced training dataset are $64\times64$ for LrMS, PAN, and GT, and $16\times16$ for MS, respectively. The resolutions of the reduced test dataset are $256\times256$ for LrMS, PAN and GT, $64\times64$ for MS, respectively. As for the full-resolution dataset, only test set is needed. The resolution is 512 for LrMS, PAN, and GT, and 128 for MS.



For the hyperspectral dataset, we crop 20 selected images into 3920 overlapping patches for training, where the sizes of HSI and MSI are $16\times 16\times 31$ and $64\times 64\times 3$, respectively.

\subsection{Benchmark}
\begin{figure*}[!ht]
    \centering
    \includegraphics[width=\linewidth]{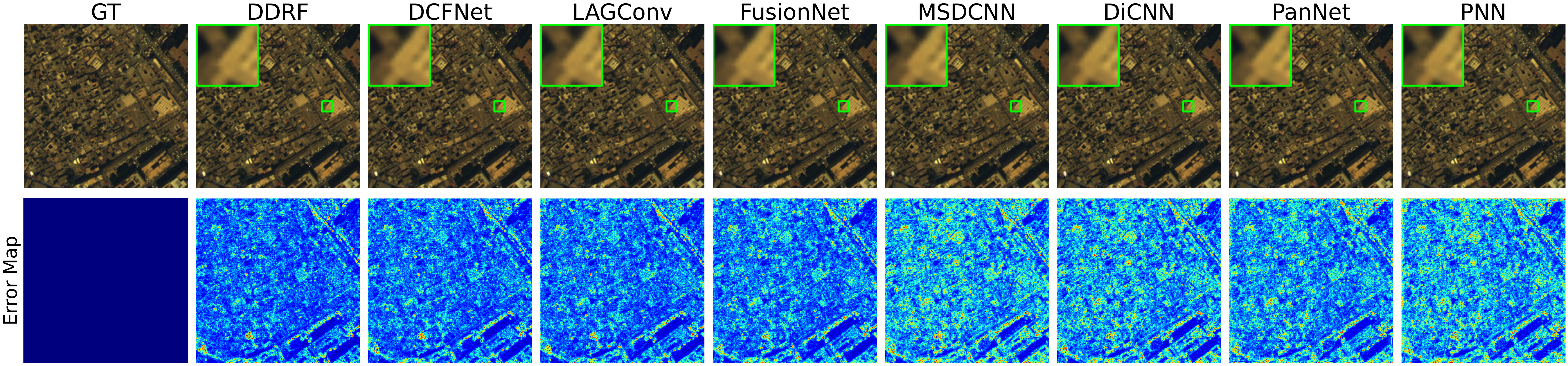}
    \caption{Visual comparisons with previous pansharpening methods on World-View3 dataset. The first row shows the results by using the pseudo-color. The second row shows the error maps between the GT and fused images. Some close-ups are depicted in green rectangles. Deeper color in the error map means better performance.}
    \label{fig:wv3_comp}
\end{figure*}

\begin{figure*}[!ht]
    \centering
    \includegraphics[width=\linewidth]{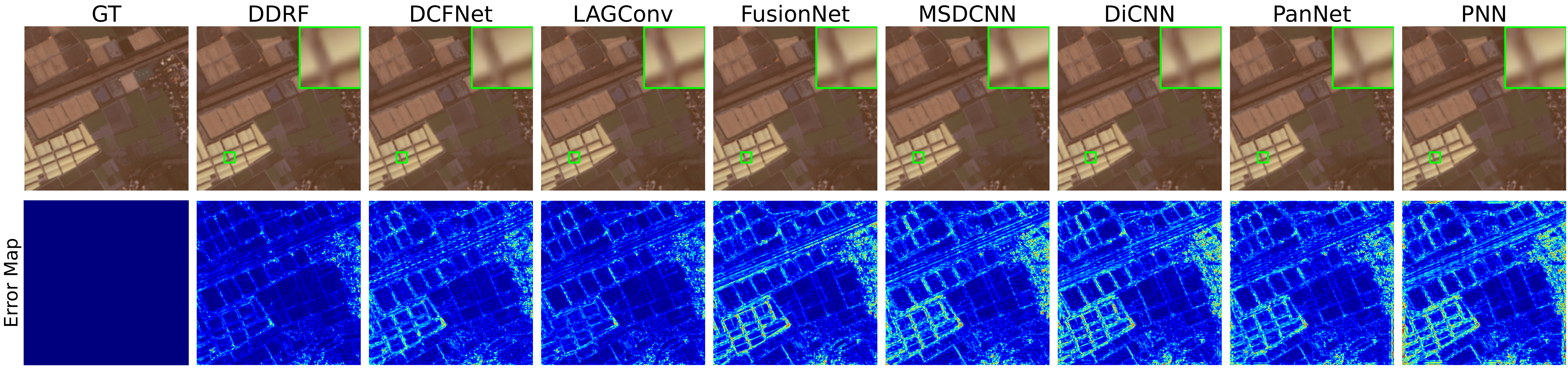}
    \caption{Visual comparisons with previous pansharpening methods on Gao-Fen2 dataset. The first row shows the results by using the pseudo-color. The second row shows the error maps between the GT and fused images. Some close-ups are depicted in green rectangles. Deeper color in the error map means better performance.}
    \label{fig:gf2_comp}
\end{figure*}

\begin{figure*}[!ht]
    \centering
    \includegraphics[width=\linewidth]{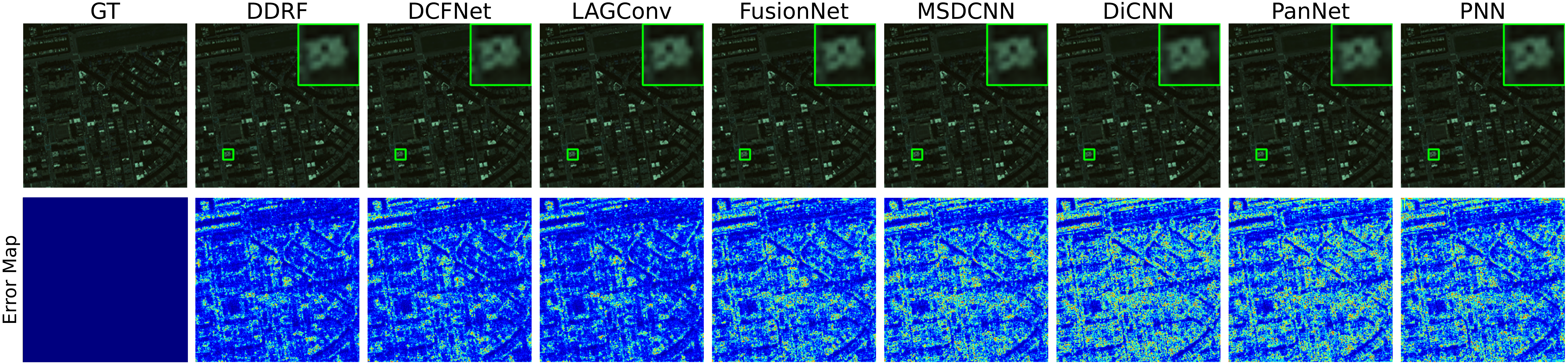}
    \caption{Visual comparisons with previous pansharpening methods on Quick-Bird dataset. The first row shows the results by using the pseudo-color. The second row shows the error maps between the GT and fused images. Some close-ups are depicted in green rectangles. Deeper color in the error map means better performance.}
    \label{fig:qb_comp}
\end{figure*}

To assess the performance of our DDRF, we compare it with various state-of-the-art methods of pansharpening (on WV3, GF2, and QB datasets) and additional MHIF (on CAVE dataset). 

Specifically, on pansharpening task, we choose some conventional methods containing optimized Brovey transform with haze correction method (BT-H)~\cite{BT-H}, band-dependent spatial-detail with physical constraints approach (BDSD-PC)~\cite{bdsd-pc}, generalized laplacian pyramid (GLP) with MTF-matched filters with a full scale
(FS) regression-based injection model (MTF-GLP-FS)~\cite{mtf-glp-fs} and a set of some competitive deep learning methods including PNN~\cite{pnn}, PanNet~\cite{yang2017pannet}, DiCNN~\cite{dicnn}, MSDCNN~\cite{msdcnn}, FusionNet~\cite{deng2020detail}, LAGNet~\cite{jin2022lagconv} and DCFNet~\cite{dcfnet}.

On the hyperspectral image fusion task, we also choose conventional and deep learning methods for comparisons, where conventional methods including the coupled sparse tensor factorization (CSTF) method~\cite{CSTF}, the fast fusion of multi-band images based on solving a Sylvester equation approach (FUSE)~\cite{FUSE}, the modulation transfer function matched generalized laplacian pyramid hyper-sharpening (MTF-GLP-HS)~\cite{mtf-glp-fs}, iterative regularization method based on tensor subspace representation method (IR-TenSR)~\cite{IR-TenSR}, low tensor-train rank-based (LTTR) method\cite{LTTR} and subspace-based low tensor multi-rank regularization method (LTMR)~\cite{LTMR}. For deep learning methods, we perform a comparison with SSRNet~\cite{SSRNet}, ResTFNet~\cite{ResTFNet}, FusFormer~\cite{hu2022fusformer}, HSRNet~\cite{HSRNet} and recent state-of-the-art DHIF~\cite{DHIF} method. All deep learning methods are trained with the sample input pairs and convergenced for fair comparisons.

\subsection{Quality Metric}
For the reduced data in pansharpening tasks, we use spectral angle mapper (SAM)~\cite{sam}, the relative dimensionless global error in synthesis (ERGAS)~\cite{ergas}, the universal image quality index (Q$2^n$)~\cite{q2n}, and the spatial correlation coefficient (SCC)~\cite{scc} as fusion metrics. The ideal values for Q$2^n$ and SCC are 1, and for SAM and ERGAS are 0 instead. The metric SAM and ERGAS are defined as Eq. \ref{eq: sam} and Eq. \ref{eq: ergas}.
Since these metrics require ground truth, they are considered reference metrics. For the full data, since there is no ground truth available, we use non-reference metrics to validate the accuracy of our DDRF. These metrics include $D_\lambda$, $D_s$, and QNR~\cite{QNR}. The QNR has an ideal value of 1 but $D_\lambda, D_s$ has ideal values of 0.

For hyperspectral data, we additionally use two metrics for evaluation, which are peak signal-to-noise ratio (PSNR) and structural similarity index measure (SSIM)~\cite{ssim}. The optimal values for PSNR and SSIM are $+\infty$ and 1. The metric PSNR is defined as Eq. \ref{eq: psnr1}.

\begin{equation} \label{eq: sam}
\begin{aligned}
    \mathrm{SAM}(\mathbf x, \mathbf{\hat x})=\frac 1 {HW}\sum_{i=1}^{HW}\cos^{-1}\left(\frac{\mathbf x_i^\top \mathbf{\hat x}_i}{\lVert \mathbf x_i\rVert_2 \lVert \mathbf{\hat x_i}\rVert_2}\right),
\end{aligned}
\end{equation}

\begin{equation} \label{eq: ergas}
\begin{aligned}
    \mathrm{ERGAS}(\mathbf x, \mathbf{\hat x})=\frac {100} c \sqrt{\frac 1 C \sum_{i=1}^C\frac{\mathrm{MSE}(\mathbf x^i, \mathbf{\hat x}^i)}{\mu_{\mathbf{\hat x^i}}^2}},
\end{aligned}
\end{equation}
\begin{equation} \label{eq: psnr1}
\begin{aligned}
\mathrm{PSNR}(\mathbf x, \mathbf {\hat x}) & = \frac 1 C \sum_{i=i}^C \mathrm{PSNR}(\mathbf x^i, \mathbf{\hat x}^i)\\
& =20\cdot\log_{10}\left(\frac{\max(\mathbf x^i)}{\sqrt{\mathrm{MSE}(\mathbf x^i, \mathbf{\hat{x}}^i)}}\right).
\end{aligned}
\end{equation}
\subsection{Results on World-View3 Dataset}
We conducted experiments to evaluate our DDRF on the World-View3 dataset and test performances on 20 testing images. We compared our method with three traditional methods and some competitive deep learning methods. The reduced assessments and full assessments are shown in Table. \ref{tab: wv3_reduced_full} To clearly demonstrate the advantages of our method, we compare it with other deep learning methods in Figure \ref{fig:wv3_comp} and zoom in on a certain position to better show the details. Additionally, the error map obtained by comparing it with the GT is also shown.

It can be seen that the deep learning methods have significantly better performances over the traditional methods on average metrics.
Among the deep learning methods, our method achieves the best performance on the reduced dataset and competitive performance on the full dataset. Our DDRF can reach 2.77 SAM and 2.05 ERGAS metrics on the reduced dataset, which outperforms all previous deep learning methods.
The error map also indicates that the image sampled by DDRF is closer to the GT (with deeper color). The full resolution image sampling by DDRF on the WV3 full dataset has similar image quality to the images fused by other non-generative methods, without color shift and spatial or spectral distortion, which indicates that our DDRF is competitive.

\begin{table*}[!ht]
    \centering 
    \caption{Result of quantitative metrics on WV3 reduced and full dataset.  Some conventional methods(the first 3 rows),  and DL-based methods are compared. The best results are in boldface and the second results are in underlining.}
    \begin{tabular}{l|cccc|ccc}
        \toprule
        \multirow{2}{*}{method}& \multicolumn{4}{c|}{Reduced} & \multicolumn{3}{c}{Full} \\  & SAM($\pm$ std) & ERGAS($\pm$ std) & Q4($\pm$ std) & SCC($\pm$ std) & $D_\lambda$($\pm$ std) & $D_s$($\pm$ std) & QNR($\pm$ std) \\
        \midrule
        BDSD-PC         & 5.4675$\pm$1.7185 & 4.6549$\pm$1.4667 & 0.8117$\pm$0.1063 & 0.9049$\pm$0.0419  & 0.0625$\pm$0.0235 & 0.0730$\pm$0.0356 & 0.8698$\pm$0.0531 \\  
        MTF-GLP-FS      & 5.3233$\pm$1.6548 & 4.6452$\pm$1.4441 & 0.8177$\pm$0.1014 & 0.8984$\pm$0.0466 & 0.0206$\pm$0.0082	& 0.0630$\pm$0.0284 & 0.9180$\pm$0.0346 \\ 
        BT-H            & 4.8985$\pm$1.3028 & 4.5150$\pm$1.3315 & 0.8182$\pm$0.1019 & 0.9240$\pm$0.0243 & 0.0574$\pm$0.0232 & 0.0810$\pm$0.0374 & 0.8670$\pm$0.0540 \\
        \midrule
        PNN             & 3.6798$\pm$0.7625 & 2.6819$\pm$0.6475 & 0.8929$\pm$0.0923 & 0.9761$\pm$0.0075 & \ul{0.0213$\pm$0.0080}	& 0.0428$\pm$0.0147 & 0.9369$\pm$0.0212 \\
        PanNet          & 3.6156$\pm$0.7665 & 2.6660$\pm$0.6887 & 0.8906$\pm$0.0934 & 0.9757$\pm$0.0088 & \textbf{0.0165$\pm$0.0074}	& 0.0470$\pm$0.0213	& \ul{0.9374$\pm$0.0271} \\
        DiCNN         & 3.5929$\pm$0.7623 & 2.6733$\pm$0.6627 & 0.9004$\pm$0.0871 & 0.9763$\pm$0.0072 & 0.0362$\pm$0.0111	& 0.0462$\pm$0.0175  & 0.9195$\pm$0.0258 \\
        MSDCNN          & 3.7773$\pm$0.8032 & 2.7608$\pm$0.6884 & 0.8900$\pm$0.0900 & 0.9741$\pm$0.0076 & 0.0230$\pm$0.0091	& 0.0467$\pm$0.0199	& 0.9316$\pm$0.0271 \\
        FusionNet       & 3.3252$\pm$0.6978 & 2.4666$\pm$0.6446 & 0.9044$\pm$0.0904 & 0.9807$\pm$0.0069 & 0.0239$\pm$0.0090	& \textbf{0.0364$\pm$0.0137}	&\textbf{0.9406$\pm$0.0197} \\
        LAGNet          & 3.1042$\pm$0.5585 & 2.2999$\pm$0.6128 & 0.9098$\pm$0.0907 & 0.9838$\pm$0.0068 & 0.0368$\pm$0.0148	& \ul{0.0418$\pm$0.0152}	&0.9230$\pm$0.0247 \\
        DCFNet          & \ul{3.0264$\pm$0.7397} & \ul{2.1588$\pm$0.4563} & \ul{0.9051$\pm$0.0881} & \ul{0.9861$\pm$0.0038} & 0.0781$\pm$0.0812 & 0.0508$\pm$0.0342 & 0.8771$\pm$0.1005 \\ \midrule
        DDRF(ours)  & \textbf{2.7722$\pm$0.5112} & \textbf{2.0484$\pm$0.4500} & \textbf{0.9191$\pm$0.0824} & \textbf{0.9877$\pm$0.0033} & 0.0741$\pm$0.0628 & 0.0586$\pm$0.0282 & 0.8715$\pm$0.0630  \\ \midrule
        Ideal value    & \textbf{0}   & \textbf{0}  & \textbf{1}  & \textbf{1}  & \textbf{0}   & \textbf{0}  & \textbf{1} \\
        \bottomrule
    \end{tabular}
    \label{tab: wv3_reduced_full}
\end{table*}

\subsection{Results on GaoFen2 Dataset}
We test our DDRF on 20 test images of Gao-Fen2 dataset. As shown in Table. \ref{tab: gf2_reduced_full}, our DDRF outperforms previous methods in nearly all reference and non-reference metrics on reduced and full datasets, achieving state-of-the-art performance. Specifically, we observed an improvement of $\approx 18\%/17\%/0.5\%$ in SAM/ERGAS/Q4 metrics when compared with the second best method, i.e., LAGNet~\cite{jin2022lagconv}. Compared with the third best method, i.e., DCFNet~\cite{dcfnet}, our DDRF improves $\approx 28\%/29\%/1.3\%$ in SAM/ERGAS/Q4 metrics. The state-of-the-art performance in the non-reference metrics on the full dataset reflects that our DDRF can also achieve good results on full-resolution images while only training on small patches.
Fig. \ref{fig:gf2_comp} shows the related close-ups in the rectangular boxes and error maps among the compared deep learning methods. The colors and edge details of object in the figure are closer to the GT. Moreover, the error maps illustrate the residual between the outcome of DDRF and the GT is visually minimal since the overall pseudo-color is the darkest.

\begin{table*}[!ht]
    \centering 
    \caption{Result of quantitative metrics on GF2 reduced and full dataset.  Some conventional methods(the first 3 rows),  and DL-based methods are compared. The best results are in boldface and the second results are in underlining.}
    \begin{tabular}{l|cccc|ccc}
        \toprule
        \multirow{2}{*}{method}& \multicolumn{4}{c|}{Reduced} & \multicolumn{3}{c}{Full} \\  & SAM($\pm$ std) & ERGAS($\pm$ std) & Q4($\pm$ std) & SCC($\pm$ std) & $D_\lambda$($\pm$ std) & $D_s$($\pm$ std) & QNR($\pm$ std) \\
        \midrule
        BDSD-PC         & 1.7110$\pm$0.3210 & 1.7025$\pm$0.4056 & 0.9932$\pm$0.0308 & 0.9448$\pm$0.0166  &0.0759$\pm$0.0301	&0.1548$\pm$0.0280	&0.7812$\pm$0.0409 \\
        MTF-GLP-FS     & 1.6757$\pm$0.3457 & 1.6023$\pm$0.3545 & 0.8914$\pm$0.0256 & 0.9390$\pm$0.0197 &0.0759$\pm$0.0301	&0.1548$\pm$0.0280	&0.7812$\pm$0.0409 \\
        BT-H            & 1.6810$\pm$0.3168 & 1.5524$\pm$0.3642 & 0.9089$\pm$0.0292 & 0.9508$\pm$0.0150  &0.0602$\pm$0.0252	&0.1313$\pm$0.0193	&0.8165$\pm$0.0305 \\
        \midrule
        PNN            & 1.0477$\pm$0.2264 & 1.0572$\pm$0.2355 & 0.9604$\pm$0.0100   & 0.9772$\pm$0.0054  &0.0367$\pm$0.0291	&0.0943$\pm$0.0224	&0.8726$\pm$0.0373 \\
        PanNet          & 0.9967$\pm$0.2119 & 0.9192$\pm$0.1906 & 0.9671$\pm$0.0099   & 0.9829$\pm$0.0035 &\ul{0.0206$\pm$0.0112}	&0.0799$\pm$0.0178	&0.9011$\pm$0.0203 \\
        DiCNN          & 1.0525$\pm$0.2310 & 1.0812$\pm$0.2510 & 0.9594$\pm$0.0101    & 0.9771$\pm$0.0058 &0.0413$\pm$0.0128	&0.0992$\pm$0.0131	&0.8636$\pm$0.0165 \\
        MSDCNN         & 1.0472$\pm$0.2210 & 1.0413$\pm$0.2309 & 0.9612$\pm$0.0108    & 0.9782$\pm$0.0050 &0.0269$\pm$0.0131	&0.0730$\pm$0.0093	&0.9020$\pm$0.0128 \\
        FusionNet      &  0.9735$\pm$0.2117    & 0.9878$\pm$0.2222   & 0.9641$\pm$0.0093 & 0.9806$\pm$0.0049 &0.0400$\pm$0.0126	&0.1013$\pm$0.0134	&0.8628$\pm$0.0184 \\
        LAGNet          & \ul{0.7859$\pm$0.1478}     & \ul{0.6869$\pm$0.1125}    & \ul{0.9804$\pm$0.0085} & \textbf{0.9906$\pm$0.0019} &0.0324$\pm$0.0130	&0.0792$\pm$0.0136	&0.8910$\pm$0.0204 \\
        DCFNet          & 0.8896$\pm$0.1577     & 0.8061$\pm$0.1369   & 0.9727$\pm$0.0100 & 0.9853$\pm$0.0024 & 0.0234$\pm$0.0116 & \ul{0.0659$\pm$0.0096} & \ul{0.9122$\pm$0.0119} \\ \midrule
        DDRF(ours)  & \textbf{0.6408$\pm$0.1203} & \textbf{0.5668$\pm$0.1010} & \textbf{0.9855$\pm$0.0078} & \ul{0.9859$\pm$0.0035} & \textbf{0.0201$\pm$0.0109} & \textbf{0.0408$\pm$0.0103} & \textbf{0.9398$\pm$0.0137} \\ \midrule
        Ideal value    & \textbf{0}   & \textbf{0}  & \textbf{1}  & \textbf{1}  & \textbf{0}   & \textbf{0}  & \textbf{1} \\
        \bottomrule
    \end{tabular}
    \label{tab: gf2_reduced_full}
\end{table*}

\subsection{Results on QuickBird Dataset}

We also conduct experiments on the QB dataset and assess the reduced and full performances of DDRF. Similarly, the reference and non-reference metrics are obtained on 20 testing images, randomly selected from QB dataset. Performances comparisons are reported in Table. \ref{tab: qb_reduced_full}.

\begin{table*}[!ht]
    \centering 
    \caption{Result of quantitative metrics on QB reduced and full dataset.  Some conventional methods(the first 3 rows),  and DL-based methods are compared. The best results are in boldface and the second results are in underlining.}
    \begin{tabular}{l|cccc|ccc}
        \toprule
        \multirow{2}{*}{method}& \multicolumn{4}{c|}{Reduced} & \multicolumn{3}{c}{Full} \\  & SAM($\pm$ std) & ERGAS($\pm$ std) & Q4($\pm$ std) & SCC($\pm$ std) & $D_\lambda$($\pm$ std) & $D_s$($\pm$ std) & QNR($\pm$ std) \\
        \midrule
        BDSD-PC         & 8.2620$\pm$2.0497 & 7.5420$\pm$0.8138 & 0.8323$\pm$0.1013 & 0.9030$\pm$0.0181 & 0.1975$\pm$0.0334	&0.1636$\pm$0.0483	&0.6722$\pm$0.0577   \\  
        MTF-GLP-FS      & 8.1131$\pm$1.9553 & 7.5102$\pm$0.7926 & 0.8296$\pm$0.0905 & 0.8998$\pm$0.0196 & 0.0489$\pm$0.0149	&0.1383$\pm$0.0238	&0.8199$\pm$0.0340 \\ 
        BT-H            & 7.1943$\pm$1.5523 & 7.4008$\pm$0.8378 & 0.8326$\pm$0.0880 & 0.9156$\pm$0.0152 & 0.2300$\pm$0.0718	&0.1648$\pm$0.0167	&0.6434$\pm$0.0645   \\ \midrule
        PNN             & 5.2054$\pm$0.9625 & 4.4722$\pm$0.3734 & 0.9180$\pm$0.0938 & 0.9711$\pm$0.0123 & 0.0569$\pm$0.0112	&0.0624$\pm$0.0239	&\ul{0.8844$\pm$0.0304}  \\
        PanNet          & 5.7909$\pm$1.1839 & 5.8629$\pm$0.8883 & 0.8850$\pm$0.0917 & 0.9485$\pm$0.0170 & \textbf{0.0410$\pm$0.0108}	&0.1137$\pm$0.0323	&0.8502$\pm$0.0390  \\
        DiCNN           & 5.3795$\pm$1.0266 & 5.1354$\pm$0.4876 & 0.9042$\pm$0.0942 & 0.9621$\pm$0.0133  & 0.0920$\pm$0.0143	&0.1067$\pm$0.0210	&0.8114$\pm$0.0310 \\
        FusionNet       & 4.9226$\pm$0.9077 & 4.1594$\pm$0.3212 & 0.9252$\pm$0.0902 & 0.9755$\pm$0.0104 & 0.0586$\pm$0.0189	&\textbf{0.0522$\pm$0.0088}	&\textbf{0.8922$\pm$0.0219} \\
        LAGNet          & 4.5473$\pm$0.8296 & \ul{3.8259$\pm$0.4196} & \ul{0.9335$\pm$0.0878} & \ul{0.9807$\pm$0.0091} &0.0844$\pm$0.0238	&0.0676$\pm$0.0136	&0.8536$\pm$0.0178 \\
        DCFNet          & \ul{4.5383$\pm$0.7397} & 3.8315$\pm$0.2915 & 0.9325$\pm$0.0903 & 0.9741$\pm$0.0101 & \ul{0.0454$\pm$0.0147} & 0.1239$\pm$0.0269 & 0.8360$\pm$0.0158 \\ \midrule
        DDRF(ours)  & \textbf{4.3529$\pm$0.7319} & \textbf{3.5659$\pm$0.2845} & \textbf{0.9380$\pm$0.0896} & \textbf{0.9838$\pm$0.0073} &0.1025$\pm$0.0266 &\ul{0.0552$\pm$0.0228} &0.8484$\pm$0.0416 \\ \midrule
        Ideal value    & \textbf{0}   & \textbf{0}  & \textbf{1}  & \textbf{1}  & \textbf{0}   & \textbf{0}  & \textbf{1} \\
        \bottomrule
    \end{tabular}
    \label{tab: qb_reduced_full}
\end{table*}

\subsection{Generalization Ability on World-View2 Dataset}
Due to the powerful fitting ability of neural networks, they often perform well on data in the same domain. However, once the test data is shifted to a domain that the network has not seen during the training process, the network usually performs poorly.
To test the generalization ability of the network, we evaluated our DDRF model trained on the WV3 dataset to test on the WV2 dataset in a zero-shot manner. Since the imaging parameters of WV2 are similar to WV3, testing on the WV2 dataset can better demonstrate the generalization performance. We also compared our method with other deep learning methods as shown in Table. \ref{tab: wv2_reduced_full}. Our method also achieves competitive results, which demonstrates the good generalization of the diffusion model. This is consistent with some works that apply the diffusion model in a zero-shot manner to tasks such as image generation~\cite{ddnm}, image classification~\cite{diffusion_zero_shot_cls}, semantic segmentation~\cite{diffumask}, and so on.

\begin{table*}[!ht]
    \centering 
    \caption{Generalization ability of DL-based methods are compared. The best results are in boldface and the second results are in underlining.}
    \begin{tabular}{l|cccc|ccc}
        \toprule
        \multirow{2}{*}{method}& \multicolumn{4}{c|}{Reduced} & \multicolumn{3}{c}{Full} \\  & SAM($\pm$ std) & ERGAS($\pm$ std) & Q4($\pm$ std) & SCC($\pm$ std) & $D_\lambda$($\pm$ std) & $D_s$($\pm$ std) & QNR($\pm$ std) \\
        \midrule
        PNN &7.1158$\pm$1.6812& 5.6152$\pm$0.9431& 0.7619$\pm$0.0928&0.8782$\pm$0.0175& 0.1484$\pm$0.0957      & 0.0771$\pm$0.0169 & 0.7869$\pm$0.0959 \\
     PanNet &\ul{5.4948$\pm$0.7126} & \textbf{4.3371$\pm$0.5197}& \textbf{0.8401$\pm$0.0796}& \textbf{0.9177$\pm$0.0081}& \ul{0.0317$\pm$0.0173}      & \textbf{0.0191$\pm$0.0123} & \textbf{0.9500$\pm$0.0275} \\
     DiCNN &6.9216$\pm$0.7898 &6.2507$\pm$0.5745 &0.7205$\pm$0.0746 &0.8552$\pm$0.0289& 0.1412$\pm$0.0661      & 0.1023$\pm$0.0195 & 0.7700$\pm$0.0505 \\
     MSDCNN &6.0064$\pm$0.6377 &4.7438$\pm$0.4939 &0.8241$\pm$0.0799 &0.8972$\pm$0.0109 & 0.0589$\pm$0.0421      & 0.0290$\pm$0.0138 & 0.9143$\pm$0.0516 \\
     FusionNet &6.4257$\pm$0.8602 &5.1363$\pm$0.5151 &0.7961$\pm$0.0737 & 0.8746$\pm$0.0134& 0.0519$\pm$0.0292      & 0.0559$\pm$0.0146 & 0.8948$\pm$0.0187 \\
     LAGNet &6.9545$\pm$0.4739 &5.3262$\pm$0.3185 &0.8054$\pm$0.0837 &0.9125$\pm$0.0101 & 0.1302$\pm$0.0856      & 0.0547$\pm$0.0159 & 0.8229$\pm$0.0884 \\
     DCFNet &5.6194$\pm$0.6039 &\ul{4.4887$\pm$0.3764} &\ul{0.8292$\pm$0.0815} &\ul{0.9154$\pm$0.0083} & 0.0649$\pm$0.0357 & 0.0700$\pm$0.0219 & 0.8690$\pm$0.0233 \\
     \midrule
     DDRF(ours) &\textbf{5.3827$\pm$0.5737} &4.6712$\pm$0.4155 &0.8217$\pm$0.0777 &0.8993$\pm$0.0129 & \textbf{0.0313$\pm$0.0376} & \ul{0.0312$\pm$0.0111} & \ul{0.9388$\pm$0.0453} \\
     \midrule
        Ideal value    & \textbf{0}   & \textbf{0}  & \textbf{1}  & \textbf{1}  & \textbf{0}   & \textbf{0}  & \textbf{1} \\
        \bottomrule
    \end{tabular}
    \label{tab: wv2_reduced_full}
\end{table*}

\subsection{Additional Experiments on Hyperspectral Dataset}
\begin{figure*}[!ht]
    \centering
    \includegraphics[width=\linewidth]{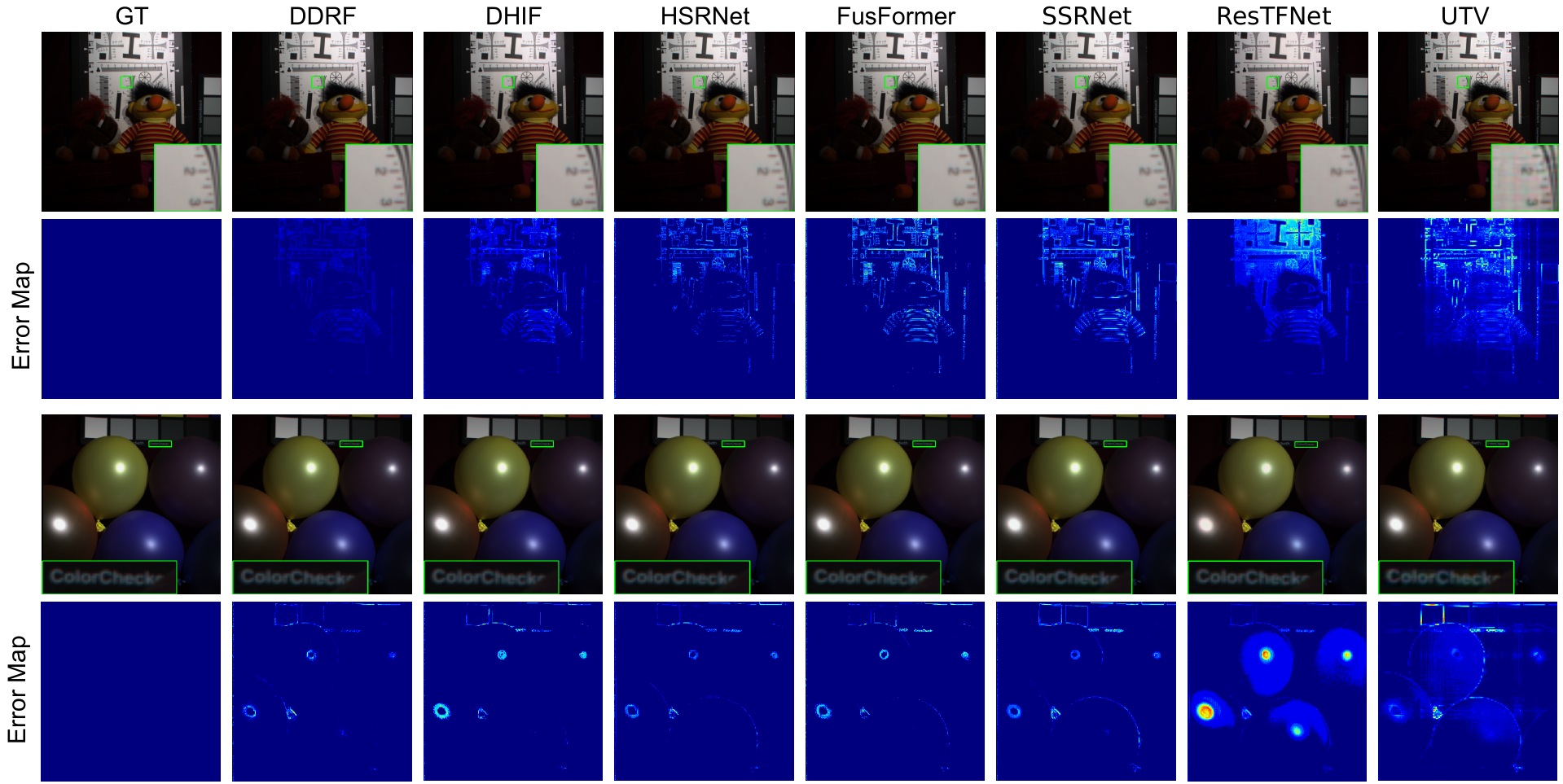}
    \caption{The first and third rows shows the results by using RGB color on "\textit{chart and stuffed toy}" and "\textit{ballons}", respectively. Some details are depicted in the green rectangles and zoomed in. The second and fourth rows show the error maps between the GT and the fused images using pseudo-color. The darker error maps indicate better performances.}
    \label{fig: cave_comp_fig}
\end{figure*}

\begin{figure}
    \label{fig: spectral_vector}
    \centering
    \subfigure[]{
    \includegraphics[width=1.0\linewidth]{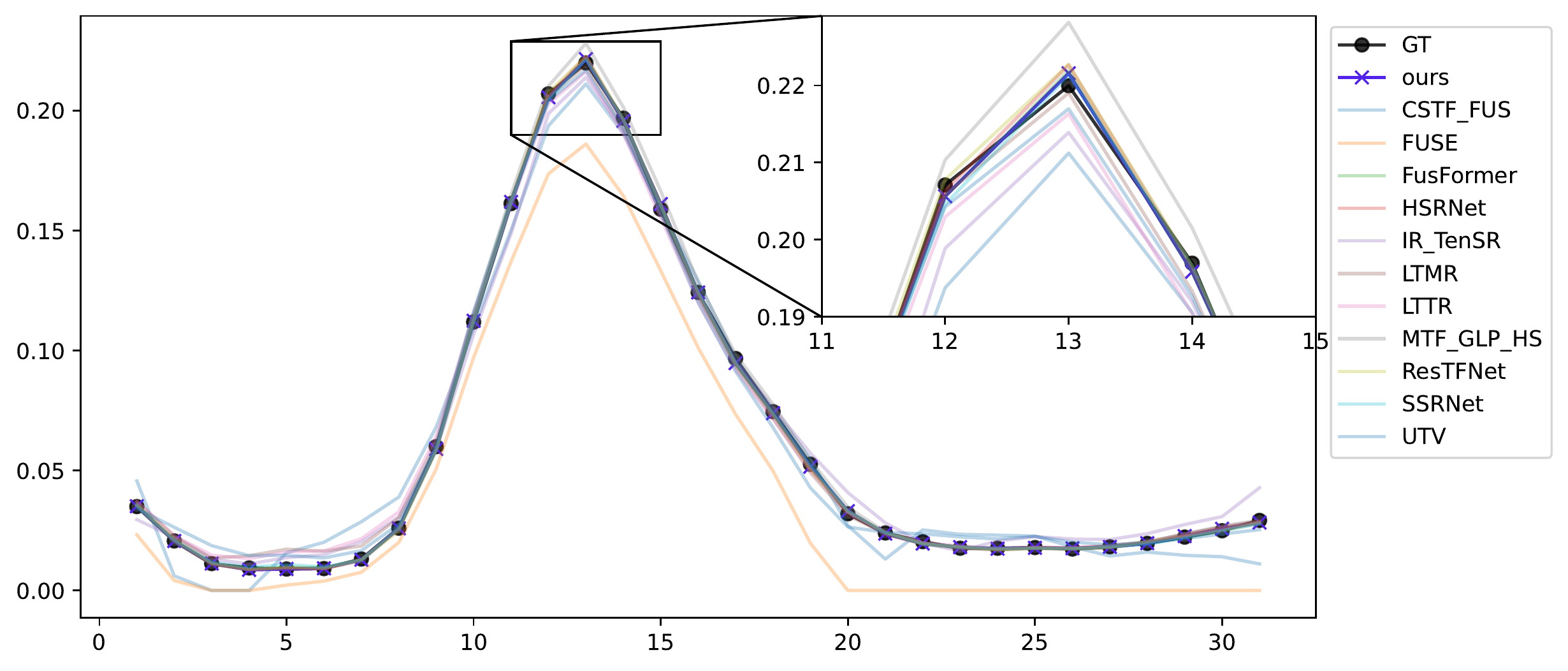}
    }

    \subfigure[]{
    \includegraphics[width=1.0\linewidth]{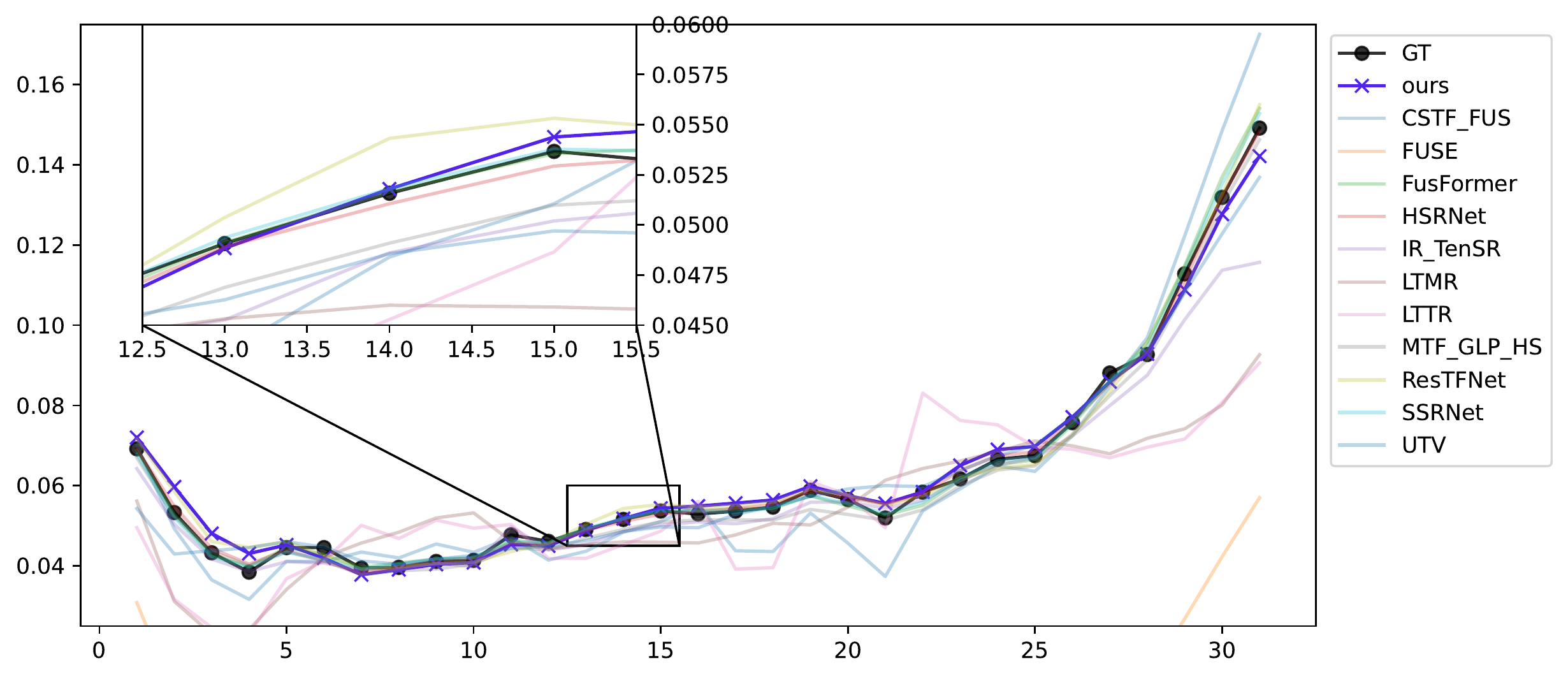}
    }
    \caption{Spectral vectors of the GT, our DDRF and benchmarks. (a) Spectral vectors in "\textit{feather}" located at position (380, 200). (b) Spectral vectors in "\textit{chart and stuffed toy"} located at position (200, 200). The horizontal axis represents the band number and the vertical axis represents pixel values.}
\end{figure}

We additionally conducted experiments on the hyperspectral CAVE dataset, and the results and comparisons with other methods can be found in Table \ref{tab: cave}. Note that compared to the state-of-the-art hyperspectral fusion method DHIF~\cite{DHIF}, our method achieves a 0.1dB higher PSNR and a higher Q2n, as well as competitive performance in other metrics. Fusion images and error maps are similarly plotted as shown in Fig. \ref{fig: cave_comp_fig}. To directly illustrate the spectral accuracy, the spectral responses of pixels in two testing images are compared and shown in Fig. \ref{fig: spectral_vector}, which shows that our DDRF has a more accurate spectral response.

\begin{figure}[!ht]
    \centering
    \includegraphics[width=\linewidth]{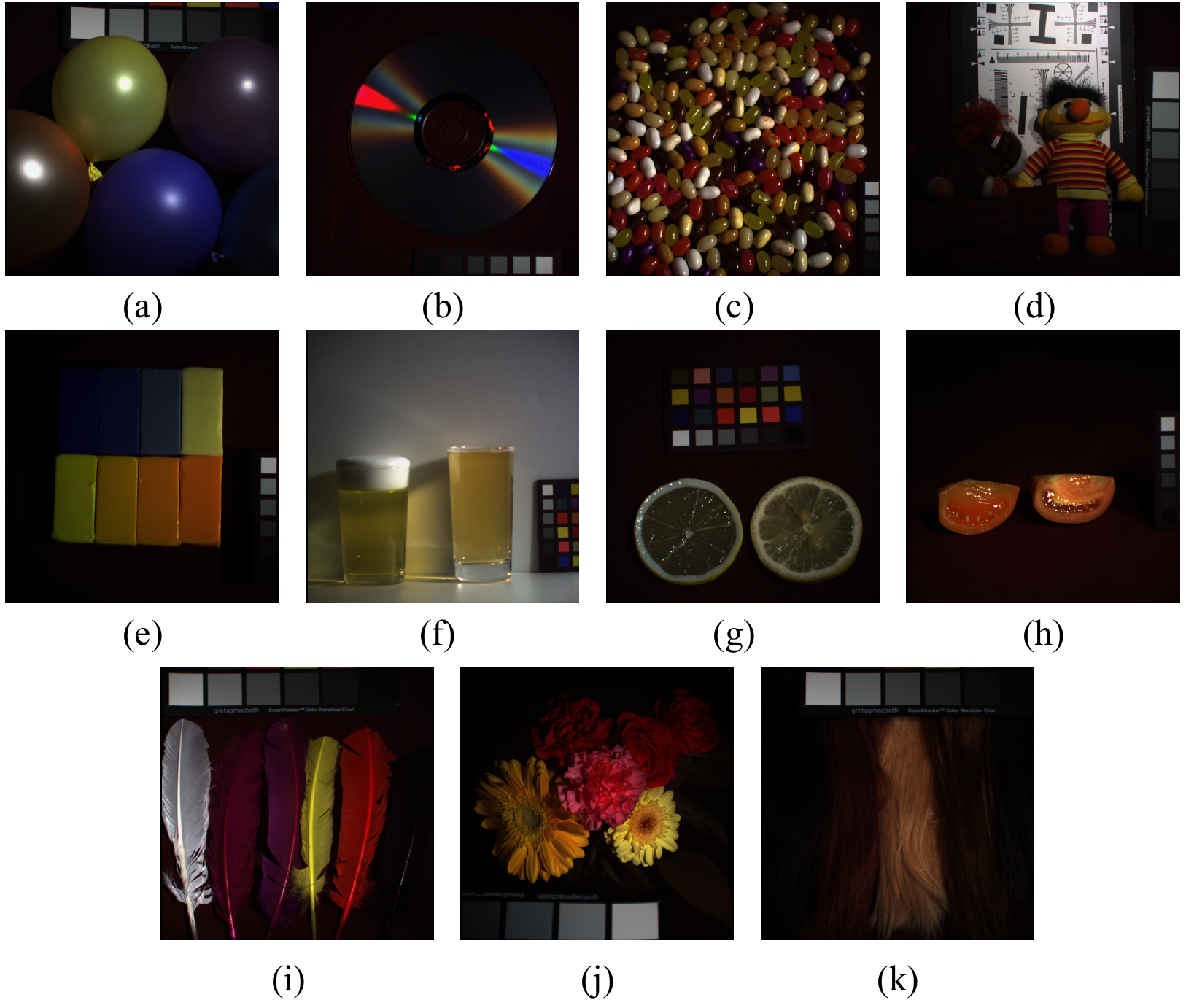}
    \caption{Testing images from the CAVE dataset. (a) \textit{Balloons}. (b) \textit{Compact
disc (CD)}. (c) \textit{Jelly beans}. (d) \textit{Chart and stuffed toy}. (e) \textit{Clay}. (f) \textit{Fake and real beers}. (g) \textit{Fake
and real lemon slices}. (g) \textit{Fake and real tomatoes}. (i) \textit{Feathers}. (j) \textit{Flowers}.
(k) \textit{Hairs}. An RGB color representation is used to depict the
images.}
    \label{fig:cave_dataset}
\end{figure}

\begin{table*}[!htbp]
\centering
\caption{Result of quantitative metrics on CAVE $\times 4$ hyperspectral dataset. Some conventional methods and deep learning methods are compared. The bold number notes the best result and the underlined number means the second-best result.}
\begin{tabular}{l|ccccc}
\toprule
method &PSNR($\pm$ std)&SSIM($\pm$ std)&Q2n($\pm$ std)&SAM($\pm$ std)&ERGAS($\pm$ std)\\ \midrule
FUSE&17.7748$\pm$4.5346 &0.3942$\pm$0.1710 &0.0892$\pm$0.0535 &34.2973$\pm$6.4079 &37.9144$\pm$7.0933\\
CSTF-FUS&34.4632$\pm$4.2806 &0.8662$\pm$0.0747 &0.6659$\pm$0.1586 &14.3683$\pm$5.3020 &8.2885$\pm$5.2848\\
IR-TenSR&35.6081$\pm$3.4461 &0.9451$\pm$0.0267 &0.7774$\pm$0.1228 &12.2950$\pm$4.6825 &5.8969$\pm$3.0455\\
LTTR&35.8505$\pm$3.4883 &0.9562$\pm$0.0288 &0.8404$\pm$0.0979 &6.9895$\pm$2.5542 &5.9904$\pm$2.9211\\
LTMR&36.5434$\pm$3.2995 &0.9631$\pm$0.0208 &0.8416$\pm$0.1031 &6.7105$\pm$2.1934 &5.3868$\pm$2.5286\\
MTF-GLP-HS&37.6920$\pm$3.8528 &0.9725$\pm$0.0158 &0.8716$\pm$0.0847 &5.3281$\pm$1.9119 &4.5749$\pm$2.6605\\
UTV&38.6153$\pm$4.0640 &0.9410$\pm$0.0434 &0.7752$\pm$0.1416 &8.6488$\pm$3.3764 &4.5189$\pm$2.8173\\
ResTFNet&45.5842$\pm$5.4647 &0.9939$\pm$0.0055 &0.9581$\pm$0.0315 &2.7643$\pm$0.6988 &2.3134$\pm$2.4377\\
SSRNet&48.6196$\pm$3.9182 &0.9954$\pm$0.0024 &0.9598$\pm$0.0309 &2.5415$\pm$0.8369 &1.6358$\pm$1.2191\\
Fusformer&49.9831$\pm$8.0965 &0.9943$\pm$0.0114 &0.9624$\pm$0.0362 &2.2033$\pm$0.8510 &2.5337$\pm$5.3052\\
HSRNet&50.3805$\pm$3.3802 &0.9970$\pm$0.0015 &0.9666$\pm$0.0290 &2.2272$\pm$0.6575 &\textbf{1.2002$\pm$0.7506}\\
DHIF & \ul{51.0721$\pm$4.1648} &\textbf{0.9973$\pm$0.0017} &0.9695$\pm$0.0267 & \bf{2.0080$\pm$0.6304} & \ul{1.2216$\pm$0.9653} \\ \midrule

DDRF(ours) & \textbf{51.1758$\pm$4.6148} & \ul{0.9971$\pm$0.0026} & \textbf{0.9737$\pm$0.0106} & \ul{2.0952$\pm$0.6471} & 1.2996$\pm$1.2822 \\
\hline
Ideal value &$\mathbf{\infty}$ &\textbf{1} &\textbf{1} &\textbf{0} &\textbf{0}\\ \bottomrule
\end{tabular}
\label{tab: cave}
\end{table*}

\subsection{Ablation Study}
We conduct an ablation study on proposed modules and techniques to verify their effectiveness.
\subsubsection{Condition Injection}
We propose two condition injection modules called style transfer modulation and wavelet modulation which are respectively responsible for coarse-grain overall style condition injection and fine-grain detail condition injection. Without using them, one direct way to inject condition information is concatenating condition and noisy input along the channel dimension. We ablate these two different techniques by training the diffusion model with the following forms,
\begin{enumerate}
    \item[a.]only concatenating condition and input together,
    \item[b.] only using coarse-grain style transfer modulation,
    \item[c.] conducting two modulations in the diffusion model.
\end{enumerate}
Each of these trained models is converged for a fair comparison. The performances on WV3 reduced dataset are reported in Table. \ref{ablation: modulations}. With the addition of style modulation and wavelet modulation, the performance of fusion exhibits a monotonically increasing trend. It should be noted that without these two modulations, DDRF will degrade to DDPM with residual learning, indicating that directly applying DDPM to the fusion task performs poorly.

\begin{table}[!htbp]
\centering
\caption{Ablation on two modulation modules.}
\resizebox{\linewidth}{!}{%
\begin{tabular}{@{}cc|cccc@{}}
\toprule
\makecell[c]{Style Transfer\\ Modulation} & \makecell[c]{Wavelet \\Modulation} & SAM($\pm$std)     & ERGAS($\pm$std)   & Q8($\pm$std)      & SCC($\pm$std)      \\ \midrule
            &    &  3.2851$\pm$0.6828       &  2.5501$\pm$0.6141&  0.8981$\pm$0.0904 & 0.9796$\pm$0.0067   \\
$\checkmark$               &                    & 3.1418$\pm$0.5789 & 2.3527$\pm$0.5116 & 0.8971$\pm$0.0940 & 0.9837$\pm$0.0047 \\
$\checkmark$ & $\checkmark$ &\textbf{2.7722$\pm$0.5112} & \textbf{2.0484$\pm$0.4500} & \textbf{0.9191$\pm$0.0824} & \textbf{0.9877$\pm$0.0033} \\ \bottomrule
\end{tabular}%
}
\label{ablation: modulations}
\end{table}

\subsubsection{Residual Learning} \label{sec: residual_learning}
To verify the effectiveness of residual learning, we return the input to the noisy HRMS as $\mathbf x_t$. Then we retrain the diffusion model until the model is converged and gain its performance on WV3 dataset as shown in Table. \ref{ablation: with_out_without_residual}. With residual learning, the diffusion model can output fused images closer to the GT. One hypothesis is that the residual distribution is "simpler" than the HRMS distribution, which benefits the model to learn more efficiently. A direct way to express the complexity of a distribution is to the entropy. We computed the pixel entropy of the HRMS and residual image (i.e., $\mathrm{HRMS-LrMS}$) and found that the entropy of HRMS is 6.498 bits-per-pixel (bpp) but the residual image is only 4.461 bpp, which indicates the distribution of the residual image is simpler.

\begin{table}[!htbp]
\centering
\caption{Ablation on using residual learning.}
\resizebox{\linewidth}{!}{
\begin{tabular}{c|cccc}
\toprule
\makecell{Residual\\ Learning} & SAM($\pm$std)     & ERGAS($\pm$std)  & Q8($\pm$std)      & SCC($\pm$std)     \\ \midrule
& 3.2397$\pm$0.4546 & 3.2096$\pm$1.003 & 0.9061$\pm$0.0837 & 0.9729$\pm$0.0113 \\
\checkmark        
&\textbf{2.7722$\pm$0.5112} & \textbf{2.0484$\pm$0.4500} & \textbf{0.9191$\pm$0.0824} & \textbf{0.9877$\pm$0.0033} \\
\bottomrule
\end{tabular}}
\label{ablation: with_out_without_residual}
\end{table}

\subsubsection{Prediction Objective}
The prediction objective of the model can be $\epsilon$, $\mathbf x_0$, or $\mathbf{v}$. $\epsilon$ means that the model needs to predict the added Gaussian noise, $\mathbf{x_0}$ indicates the model predicts the original clean image from the noisy image, and $\mathbf{v}$ reflects that model predicts the weighted sum of noise and $\mathbf{x_0}$. We changed the training objectives of the network to these three types separately and trained the diffusion model to convergency on the WV3 dataset. Their performances are shown in Table. \ref{tab: pred_obj}. As can be seen that predicting $\mathbf{x_0}$ performs the best. We suppose that for the small-scale datasets, $\mathbf{x_0}$ is more straightforward than $\epsilon$ and $\mathbf{v}$ since the sample density is more concentrated, but as for large-scale datasets, predicting $\epsilon$ and $\mathbf{v}$ can force the network to learn detailed denoising trajectory which is useful to generate images with high-level conditions, such as classification label~\cite{adm}, text~\cite{stable_diffusion} and bounding box~\cite{controlnet}.

\begin{table}[!htbp]
    \centering
    \caption{Ablation on training objectives of the diffusion model.}
    \resizebox{\linewidth}{!}{
    \begin{tabular}{c|cccc}
        \toprule
        Objective & SAM($\pm$std) & ERGAS($\pm$std) & Q8($\pm$std) & SCC($\pm$std) \\
        \midrule
         $\epsilon$ & 3.7702$\pm$0.6397 & 2.7954$\pm$0.6516 & 0.8388$\pm$0.1181 & 0.9794$\pm$0.0064\\
         $\mathbf{v}$ & 3.4853$\pm$0.6080 & 2.6624$\pm$0.5912 & 0.8715$\pm$0.1025 & 0.9808$\pm$0.0052 \\
         $\mathbf x_0$ & \textbf{2.7722$\pm$0.5112} & \textbf{2.0484$\pm$0.4500} & \textbf{0.9191$\pm$0.0824} & \textbf{0.9877$\pm$0.0033} \\
        \bottomrule
    \end{tabular}
    }
    \label{tab: pred_obj}
\end{table}

\section{Conclusions}
In this paper, we proposed a denoising diffusion model named DDRF for remote sensing image fusion tasks. To make DDRF suit the fusion task, we design two novel feature modulation modules: the style modulation module and wavelet modulation module responsible for coarse-grained style information modulation and fine-grained details modulation, respectively.  We also discussed residual learning and some choices of training objectives. Experiments conducted on commonly used satellite datasets and an additional hyperspectral dataset demonstrated our DDRF can outperform previous image fusion methods. Our method first introduces the denoising diffusion model into the remote sensing fusion, we hope our work can inspire later works for better applying the diffusion model for the remote sensing fusion field.

\appendices




\ifCLASSOPTIONcaptionsoff
  \newpage
\fi



\bibliographystyle{IEEEtran}
\bibliography{bibtex/bib/reference}

\end{document}